\documentclass[9pt]{article}
\usepackage{spconf,amsmath,graphicx}
\usepackage[numbers]{natbib}
\usepackage{multirow}
\usepackage{textcomp}
\usepackage{graphicx,scalerel}
\usepackage{moredefs}
\usepackage{color}
\usepackage{hyperref}
\hypersetup{colorlinks=true}
\usepackage[final]{changes}

\newcommand\wh[1]{\hstretch{2}{\hat{\hstretch{.5}{#1}}}}

\definecolor{darkspringgreen}{rgb}{0.09, 0.45, 0.27}

\title{Transformer Based Deliberation for Two-Pass Speech Recognition}
\name{Ke Hu, Ruoming Pang, Tara N. Sainath, Trevor Strohman}
\address{Google, Inc., USA \\
\fontsize{9}{9}\selectfont\ttfamily\upshape
\{huk,rpang,tsainath,strohman\}@google.com}
\begin{document}
%
\maketitle
\begin{abstract}
Interactive speech recognition systems must generate words quickly while also producing accurate results. Two-pass models excel at these requirements by employing a first-pass decoder that quickly emits words, and a second-pass decoder that requires more context but is more accurate. Previous work has established that a deliberation network can be an effective second-pass model. The model attends to two kinds of inputs at once: encoded audio frames and the hypothesis text from the first-pass model. In this work, we explore using transformer layers instead of long-short term memory (LSTM) layers for deliberation rescoring. In transformer layers, we generalize the ``encoder-decoder" attention to attend to both encoded audio and first-pass text hypotheses. The output context vectors are then combined by a merger layer.  Compared to LSTM-based deliberation, our best transformer deliberation achieves 7\% relative word error rate improvements along with a 38\% reduction in computation. We also compare against non-deliberation transformer rescoring, and find a 9\% relative improvement.
\end{abstract}
\section{Introduction}
\label{sec:intro}

End-to-end (E2E) automatic speech recognition (ASR) has made rapid progress in recent years \cite{graves2012sequence, chorowski2015attention, chan2016listen, li2020comparison, he19streaming, chiu18, sainath2020streaming}.
Representative models include streaming models such as the recurrent neural network transducer (RNN-T)~\cite{graves2012sequence}, attention-based models~\cite{bahdanau2014neural,chorowski2015attention,chan2016listen}, and transformer-based models \cite{yeh2019transformer, zhang2020transformer,lu2020exploring, tsunoo2020streaming}. 
Compared to sophisticated conventional models \cite{pundak2016lower, zhou2020rwth}, E2E models such as RNN-T and Listen, Attend and Spell (LAS) have shown competitive performance~\cite{chiu18, he19streaming,sainath2020streaming, li2020developing}. To further improve recognition accuracy, a two-pass LAS rescoring model has been proposed in~\cite{sainath2019twopass}, which uses a non-streaming LAS decoder to \emph{rescore} the RNN-T hypotheses. The rescorer attends to audio encoding from the encoder to re-rank the first-pass hypotheses. \cite{sainath2020streaming} shows that by using an RNN-T model which is capable of endpointing based on an end-of-query token, LAS rescoring can outperform a large conventional model~\cite{pundak2016lower} in both latency and word error rate (WER).

Recently, deliberation network~\cite{xia2017deliberation} has been proposed for second-pass rescoring~\cite{hu2020deliberation} and achieved state-of-the-art WER on Google Voice Search (VS). By using multi-source attention in a LAS decoder, deliberation attends to both encoder outputs and first-pass hypotheses for rescoring. For comparison, LAS rescoring only attends to encoder outputs for rescoring~\cite{sainath2019twopass, sainath2020streaming}, and neural correction models post-process hypotheses using only text hypotheses or lattice~\cite{ma2020neural,peyser2019improving,guo2019spelling}. 

While long short-term memory (LSTM) has been a popular building block for E2E models, there has been a continuing success in applying transformer models~\cite{vaswani2017attention} in ASR~\cite{wang2019transformer, lu2020exploring, zhang2020transformer, yeh2019transformer, li2020parallel, hrinchuk2020correction, li2020comparison}. Instead of using a recurrent mechanism to model temporal dynamics, the transformer uses multi-headed attention to associate sequential elements in one step.~\cite{wang2019transformer, lu2020exploring} incorporate transformer layers to conventional models for acoustic modeling. For E2E models, the transformer has been adapted or applied to streaming models~\cite{zhang2020transformer, yeh2019transformer, tsunoo2020streaming} and non-streaming models~\cite{li2020comparison}. Comparative studies~\cite{karita2019comparative, li2020comparison} show that transformer-based models outperform their recurrent neural network (RNN) counterparts in a number of tasks. \added{Transformers have also been applied in post-processing for E2E models. \cite{hrinchuk2020correction, li2020developing} use transformer for spelling correction. ~\cite{li2020parallel} applies transformer decoder in second-pass \emph{rescoring}. The non-autoregressive nature of a transformer layer enables token-level parallel rescoring, i.e., the rescoring of one token in a sequence does not depend on the previous one. This can significantly reduce latency on a matrix computation friendly hardware such as Tensor Processing Units (TPU).}

\begin{figure*}[h]
  \centering
   \includegraphics[scale=0.53]{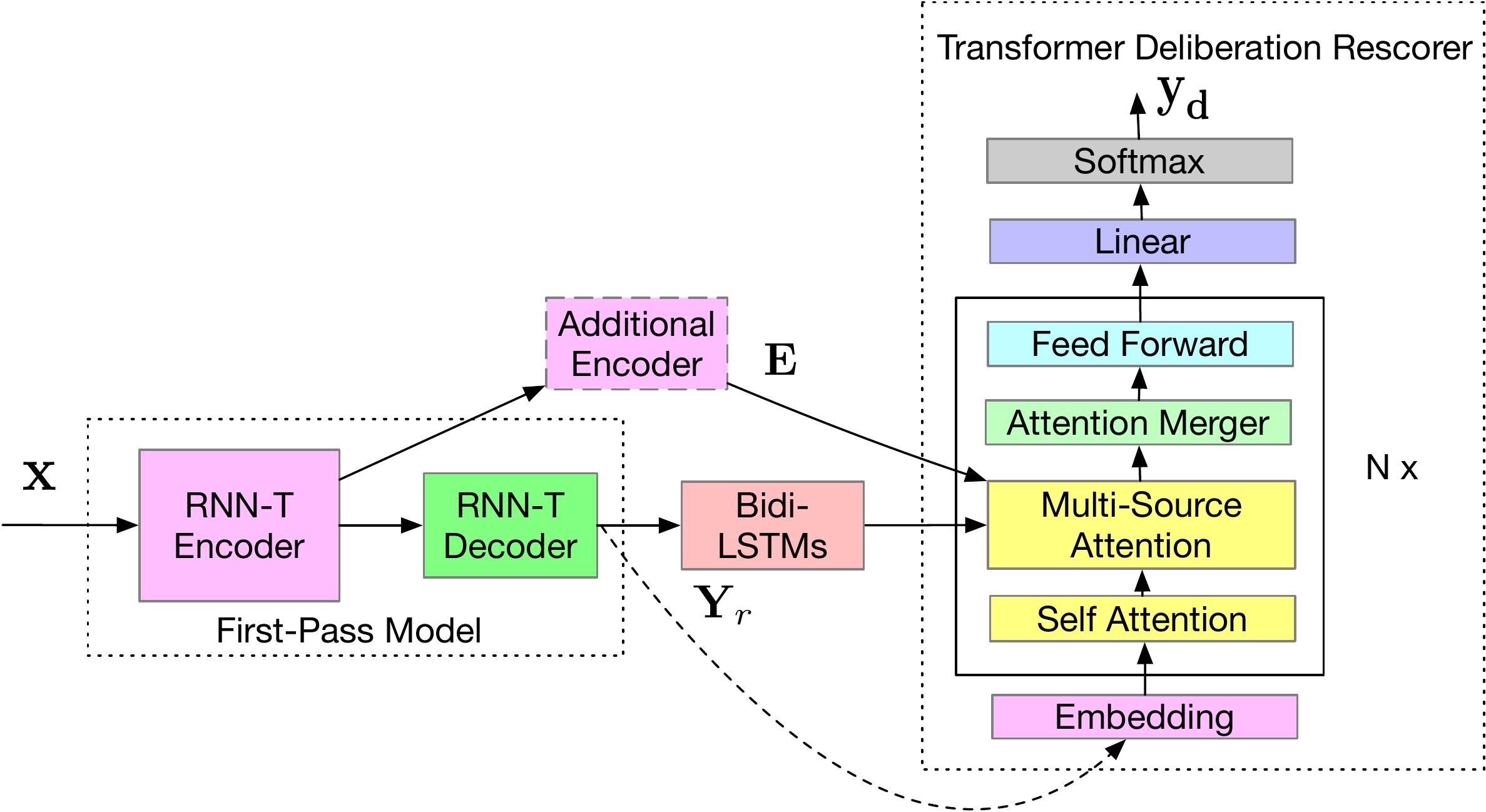}
   \caption{Diagram of a two-pass transformer deliberation model with an optional additional encoder (dashed box). The RNN-T model generates first-pass hypotheses, and a second-pass transformer deliberation rescorer re-ranks the first-pass results. The ``encoder-decoder" attention of the transformer attends to two sources: Acoustic encoding and first-pass hypothesis encoding. The two output context vectors are combined using a merger layer. In addition, first-pass hypotheses are bidirectionally encoded to extract context information.}
   \label{fig:delib}
\end{figure*}

In this work, we develop a transformer-based deliberation rescorer, which is more TPU friendly than its LSTM counterpart~\cite{hu2020deliberation}. We use a transformer decoder in the second pass rescoring, i.e., the first-pass RNN-T model generates first-pass hypotheses, and then we use a transformer decoder to \emph{rescore} the first-pass hypotheses. In rescoring, the transformer ``encoder-decoder" attention attends to two sources: Encoder outputs and first-pass hypothesis encoding. The resultant context vectors are then combined by a merger layer to produce a new context vector that has the same dimension as the output from the previous self-attention layer. Our model is different from transformer rescoring~\cite{li2020parallel} where the decoder only attends to the encoder outputs. As in~\cite{hu2020deliberation}, our model encodes first-pass hypotheses bidirectionally to model context. When encoding multiple hypotheses, we model the hypothesis order (i.e., rank in the n-best list) using a learned embedding~\cite{variani2020neural} and add that to the embedding of wordpiece tokens  in the corresponding hypothesis. We note that the wordpiece-based~\cite{schuster2012japanese} sequences are usually short for VS queries, and computation can be paralleled to improve latency.

We conduct experiments on the same training data as in~\cite{narayanan2019recognizing,hu2020deliberation,li2018multi}, which is from multiple domains including Voice Search, YouTube, Telephony such as SwitchBoard \cite{godfrey1992switchboard} and Fisher \cite{cieri2004fisher}, etc. We researched different types of merge layer, i.e., sum, concatenation, attention or gated averaging, to combine the attention vectors from encoder outputs and first-pass hypothesis encoding, and find sum is a more effective and efficient method. Similar to the previous study~\cite{hu2020deliberation}, we observe that using multiple RNN-T hypotheses, additional encoder (AE) layers, and minimum WER (MWER) training~\cite{prabhavalkar2018minimum} improves WER. Order embedding and joint training further improve WER. In summary, our best proposed transformer deliberation rescorer achieves 7\% WER reduction compared to the best LSTM-based deliberation~\cite{hu2020deliberation} for VS and proper noun recognition, and reduces computation by 38\% relative. Compared to non-deliberation transformer rescoring~\cite{li2020parallel}, our deliberation rescorer shows 9\% and 11\% relative improvement. To achieve a tradeoff between WER and computation, we explore reducing the model complexity by encoding a single hypotheses for deliberation, and achieves 4\% and 8\% WER reduction for VS and proper noun recognition, respectively, by using 1.25 times computation of~\cite{li2020parallel}.

\section{Transformer-Based Deliberation}
\label{sec:pagestyle}

\subsection{Transformer Deliberation Architecture}
\label{sec:architecture}

As shown in Fig. \ref{fig:delib}, our transformer deliberation model has two main parts: A first-pass RNN-T model, and a second-pass transformer deliberation rescorer. The overall deliberation structure resembles that of~\cite{xia2017deliberation, hu2020deliberation}. The input of the RNN-T encoder are log-mel filterbank energies, $\mathbf{X} = ( \mathbf{x}_1,...,\mathbf{x}_T )$, where $T$ denotes the number of frames. The encoder output is then fed to an RNN-T decoder to produce first-pass decoding results $\mathbf{Y}_r = ( \mathbf{y}_r^1,...,\mathbf{y}_r^B )$ in a streaming fashion, where $B$ is the beam search size. Note our RNN-T decoder is the same as~\cite{sainath2020streaming} which includes both the prediction and joint networks in~\cite{graves2012sequence}.

The second-pass model is a multi-source transformer rescorer. The general structure of the transformer rescorer is similar to~\cite{vaswani2017attention} and consists of $N$ transformer layers, each of which contains self-attention, encoder-decoder attention, and a feed forward layer. Different from~\cite{vaswani2017attention}, our ``encoder-decoder" attention attends to two sources. One is the encoder outputs, which are optionally encoded by an additional encoder (dashed box in Fig. \ref{fig:delib}) to generate audio encoding $\mathbf{E}$, and the other is encoding of first-pass hypotheses. \added{Note that to encode first-pass hypotheses $\mathbf{Y}_r$, we still use a bidirectional LSTM since the focus of this work is on the decoder.} We encode multiple hypotheses $\{\mathbf{y}_r^i\}$ separately using the same bidirectional encoder, where $i=1,...,H$, and $H \le B$ is the number of hypotheses used. Encoding of multiple hypotheses is then concatenated in time for attention computing. \added{Our previous LSTM-based deliberation model \cite{hu2020deliberation} did not consider the order of n-best hypotheses during encoding. In this work, we propose to embed the hypothesis order (i.e. hypothesis rank in the n-best list) using a learned embedding \cite{variani2020neural} and add that to the embedding of each token in the hypothesis.} We keep the additional encoder unidirectional due to latency considerations.

The multi-source ``encoder-decoder" attention generates two context vectors, and we use a merger layer to combine them. We tried multiple merging options such as concatenation, sum, attention, and gated average~\cite{bapna2019nonparametric}. In the attention case, we use the previous self-attention layer output as query and two source context vectors as key and values. For gated average, we take the concatenated source contexts as inputs and uses a softmax layer to predict weights for two sources. The output vector is then a weighted average of the two sources.

For structural comparison, the proposed model uses a transformer-based multi-source decoder for rescoring, instead of a LSTM-based decoder~\cite{hu2020deliberation}. Compared to~\cite{li2020parallel}, our model attends to both encoder outputs and first-pass hypothesis encoding, while~\cite{li2020parallel} only attends to encoder outputs for rescoring.

\subsection{Training and Rescoring}
\label{sec:train_decode}

Similar to~\cite{hu2020deliberation}, we use a two-step training process: Train the RNN-T model first, and then fix the RNN-T parameters and only train the transformer deliberation rescorer and additional encoder layers. In the second pass, for the $u$-th step, the transformer deliberation decoder is trained to predict $p(y_u|\mathbf{X}, \{\mathbf{y}_r^i\}, y_1,...,y_{u-1})$ by using encoder outputs $\mathbf{X}$, first-pass hypotheses $\{\mathbf{y}_r^i\}, i = 1,...,H$, and all previously predicted tokens using cross-entropy (CE) loss.

Minimum word error rate (MWER) training~\cite{prabhavalkar2018minimum} has been found effective in previous works~\cite{sainath2020streaming, hu2020deliberation, li2020parallel},  and thus we fine-tune the rescorer by training using the MWER loss. MWER training optimizes the expected word error rate by using n-best hypotheses:
\begin{equation}
L_{\text{MWER}}(\mathbf{x}, \mathbf{y}^*) = \sum_{i=1}^{B} \hat{P}(\mathbf{y}_d^i|\mathbf{X}, \mathbf{Y}_r)[W(\mathbf{y}_d^i, \mathbf{y}^*) - \wh{W}]
\label{eq:mwer}
\end{equation}
\noindent
$\mathbf{y}_d^i$ is the $i$-th hypothesis from the transformer deliberation decoder. To make the MWER training match rescoring, we compute $P(\mathbf{y}_d^i|\mathbf{X}, \mathbf{Y}_r)$ by rescoring the first-pass hypotheses $\mathbf{Y}_r$. In addition, $W(\mathbf{y}_d^i, \mathbf{y}^*)$ is the number of word errors for $\mathbf{y}_d^i$ w.r.t the ground truth target $\mathbf{y}^*$, and $\wh{W}$ is the average number of word errors of all n-best hypotheses. $\hat{P}(\mathbf{y}_d^i|\mathbf{X}, \mathbf{Y}_r)$ is the normalized probability of the $i$-th hypothesis over all hypotheses. $B$ is the beam size. In practice, we combine the MWER loss with CE loss to stabilize training: 
\begin{equation}
L'_{\text{MWER}}(\mathbf{x}, \mathbf{y}^*)=L_{\text{MWER}}(\mathbf{x}, \mathbf{y}^*) + \alpha L_{\text{CE}}(\mathbf{x}, \mathbf{y}^*)
\end{equation}
where $\alpha=0.01$ as in~\cite{prabhavalkar2018minimum}.

Similar to~\cite{hu2020deliberation}, we also explore jointly training the model by updating the whole RNN-T model, and the second-pass rescorer in the second step of the training. We show that this further improves the rescoring quality in Sect. \ref{subsec:comp}. 

Our overall decoding also consists of two passes: 1) The RNN-T model generates the first-pass results $\mathbf{Y}_r$, and 2) The transformer deliberation rescorer attends to both $\mathbf{Y}_r$ (or a subset) and $\mathbf{E}$ to re-rank first-pass hypotheses to produce $\mathbf{y}_d$. In rescoring, we run the deliberation decoder on $\mathbf{Y}_r$ in a teacher-forcing mode~\cite{sainath2019twopass, li2020parallel}. \replaced{A major difference}{Note that different} from~\cite{sainath2019twopass, li2020parallel} is that deliberation rescoring has access to bi-directional encoding of complete first-pass hypotheses.
\section{Experimental Setup}
\label{sec:exp}

\subsection{Model Details}
We use a domain-id RNN-T~\cite{sainath2020streaming, li2018multi} as the first-pass model. A domain-id is fed to the RNN-T encoder as a one-hot vector to differentiate 4 domains: Search, farfield, telephony and YouTube. The RNN-T encoder is a 8-layer LSTM~\cite{hochreiter1997}. Every LSTM is unidirectional with 2,048 hidden units followed by 640-dimensional projection. A time-reduction layer is added after the second LSTM to increase the inference speed without accuracy loss. The RNN-T decoder contains a prediction network with 2-layer LSTM and a joint-network with a single feed-forward layer of 640 hidden units. The RNN-T decoder is trained to predict 4,096 mixed-case wordpieces~\cite{schuster2012japanese}.

Our transformer deliberation rescorer consists of 4 transformer layers, each of which has a model dimension $d_{model} = 640$ and feed forward layer dimension $d_{ff} = 2560$, proportional to the sizes in~\cite{vaswani2017attention}. Both self-attention and ``encoder-decoder" attention use multi-headed dot-product attention with 8 heads. We have two sources for the ``encoder-decoder" attention and it is applied for all 4 layers. Bidirectional encoding of multiple RNN-T hypotheses are concatenated and used as the second source. The hypotheses are first padded with end-of-sentence label to a length of 120, and then each token in a hypothesis is embedded using the same 640-dimensional embedding layer as in the transformer decoder. Hypothesis order embedding is then added to each token embedding. The embeddings are then encoded by a 2-layer bidirectional LSTM encoder. Each LSTM has 2,048 hidden units and 320-dimensional projection. The bidirectional LSTM has a total of 32M parameters, more than that of the LSTM-based deliberation~\cite{hu2020deliberation} because we reuse 640-dimensional embedding from the transformer decoder.

The two output context vectors are then summed to produce a single 640-dimensional context vector. Sum is the most effective and efficient among all methods we explored as in Sect. \ref{ssec:merge}. In total, the rescorer has 37M parameters, including a 6.3M-parameter ``encoder-decoder" attention on RNN-T hypotheses. The rescorer has a 4,096-dimensional softmax layer to predict the same mixed-case wordpieces as the RNN-T.

\subsection{Large Scale Training}
We use the multidomain datasets described in~\cite{narayanan2019recognizing} for large-scale training. The English utterances are sampled from multiple domains such as general Google traffic, far-field environments, telephony conversations, and YouTube. These utterances are anonymized and hand-transcribed except for YouTube whose transcripts are extracted in a semi-supervised way~\cite{liao2013large}. We augment the clean training utterances by artificially corrupting them by using a room simulator, varying degrees of noise, and reverberation such that the signal-to-noise ratio (SNR) is between 0dB and 30dB~\cite{kim2017generation}. We also use mixed-bandwidth utterances at 8kHz or 16 kHz for training~\cite{yu2013feature}.

For the feature extraction front-end, we divide and window a speech waveform using 32-ms hanning windows at a rate of 10 ms, and then compute 128-dimensional log-Mel spectra. Each log-spectrum is stacked with three previous frames to form a 512-dimensional vector, and then downsampled to a 30-ms frame rate. Our models are trained in Tensorflow~\cite{abadi2016tensorflow} using the Lingvo framework~\cite{shen2019lingvo}  on 8$\times$8 TPU slices with a global batch size of 4,096. As in~\cite{sainath2019twopass}, we use constant learning rate for training and maintain an exponential moving average~\cite{polyak1992acceleration} of the trained model parameters for evaluation.

\subsection{Evaluation}
\label{subsec:eval}

Our main test set is Google Voice Search (VS), which includes \texttildelow14K anonymized hand-transcribed VS utterances sampled from Google traffic. To measure the model effectiveness on proper noun recognition, we also use a side-by-side (SxS) test set, which contains utterances where the E2E models~\cite{sainath2019twopass} performs inferior to a state-of-the-art conventional model~\cite{pundak2016lower}. A Search domain ID is used for both test sets. In addition to WER, we also report computational complexity in GIGA floating-point operations (GFLOPS) for second-pass decoding. The GFLOPS is computed as total operations needed for bidirectional encoding and second-pass rescoring, and is proportional to beam size ($B$) and output sequence length ($N$):
\begin{equation}
\texttt{GFLOPS} = \mathcal{O}(BN)
\label{eq:ops}
\end{equation}
In our experiments, we use $B=8$ and $N=12$ (wordpiece length for the VS utterance corresponding to $90$th percentile rescoring latency~\cite{li2020parallel}).

\section{Results}
\label{sec:results}
We first present ablation studies to find the best architecture for the proposed transformer deliberation rescorer and then compare to the best-performing LSTM-based deliberation rescorer~\cite{hu2020deliberation} and non-deliberation transformer rescoring~\cite{li2020parallel}. We do not use any voice-activity detector or endpointer \cite{chang2019joint} in either first-pass decoding or second-pass rescoring.


\subsection{Attention Merger}
\label{ssec:merge}

We tried 4 different ways, i.e., concatenation, sum, attention, and gated average, to merge the source context vectors in an attention merger layer as shown in Fig. \ref{fig:delib}. The output is a merged context vector which has the same dimension as the output of the previous self-attention layer. In the concatenation case, we first project each source vector to half of its dimension and then concatenate them. We find that sum and gated average performs the best: 5.7\% WER for VS. We thus choose sum for simplicity and use it for following experiments. 

\begin{table}[h]
\centering
\begin{tabular}{ |c|c|c|c|c| }
    \hline
    ID & E1 & E2 & E3 & E4 \\ \hline\hline
    Attention Merger & Concat & Sum & Atten & Gated \\ \hline
    VS WER (\%) & 5.8 & 5.7 & 6.1 & 5.7 \\ \hline
\end{tabular}
\caption{WERs (\%) of transformer deliberation rescoring by using different types of merging layers.}
\label{tab:wer_merge}
\end{table}

\subsection{Additional Encoder}
\label{ssec:ae}

Additional encoder (AE) layers (dashed box in Fig. \ref{fig:delib}) are found to improve recognition for LSTM-based deliberation~\cite{hu2020deliberation}. The AE consists of a 2-layer LSTM with 2,048 hidden units followed by 640-dimensional projection per layer. In Table \ref{tab:ae}, we observe that transformer deliberation rescoring also benefits from AE layers. A rescorer with AE layers (\texttt{E5}) performs around 4\% relatively better than without (\texttt{E4}).

\begin{table}[h]
\centering
\begin{tabular}{ |c|c|c|c| }
    \hline
    ID & Model & VS WER (\%) \\ \hline\hline
    E4 & w/o AE & 5.7 \\ \hline
    E5 & w/ AE & 5.5 \\ \hline
\end{tabular}
\caption{WERs (\%) with or without AE layers.}
\label{tab:ae}
\end{table}

\subsection{Hypothesis Encoding and MWER}
\label{ssec:mwer}

Similar to~\cite{hu2020deliberation}, we also observe improvement by using multiple first-pass hypotheses and MWER training. However, the improvement is smaller compared to our previous study~\cite{hu2020deliberation}, probably because our RNN-T baseline is significantly better (see Table \ref{tab:wer}). In addition, we find that a 4-hypothesis MWER rescorer already performs the best and thus use it for comparison in Sect. \ref{subsec:comp} considering computational efficiency. The models in Table \ref{tab:wer_num_hyp} all have AE layers. 

\begin{table}[h]
\centering
\begin{tabular}{ |c|c|c|c|c| }
    \hline
    ID & E6 & E7 & E8 & E9 \\ \hline\hline
    Model & 1 hyp & 2 hyp & 4 hyp & 8 hyp \\ \hline
    Trans. Delib. & 5.5 & 5.5 & 5.4 & 5.4 \\ \hline
    + MWER & 5.4 & 5.4 & 5.3 & 5.3 \\ \hline
\end{tabular}
\caption{VS WERs (\%) of transformer deliberation rescoring by attending to different numbers of RNN-T hypotheses and MWER training.}
\label{tab:wer_num_hyp}
\end{table}

\subsection{Hypothesis Order Embedding}
\label{ssec:order}
In this work, we also propose to embed the hypothesis order and add that to the embedding of each wordpiece token in a hypotheses. This let the model differentiate hypotheses in the first-pass n-best list. When using encoding of 4 first-pass hypotheses (\texttt{E8}), we find the order embedding slightly improves WER (Table \ref{tab:order}).

\begin{table}[h]
\centering
\begin{tabular}{ |c|c|c|c| }
    \hline
    ID & Model & VS WER (\%) \\ \hline\hline
    E8 & w/o order embedding & 5.3 \\ \hline
    E10 & w/ order embedding & 5.2 \\ \hline
\end{tabular}
\caption{WERs (\%) with or without order embedding.}
\label{tab:order}
\end{table}

\begin{table*}[h]
\centering
\begin{tabular}{ |c|l|c|c|c|c|c|c|c|c|c| }
    \hline
    \multirow{2}{*}{ID} & \multirow{2}{*}{Model} & \multirow{2}{*}{\shortstack{\# Encoded First-Pass \\ Hypotheses (H)}} & \multicolumn{2}{|c|}{WER (\%)} & \multirow{2}{*}{\shortstack{2nd Pass \\ GFLOPS}} \\ \cline{4-5}
     & & & VS & SxS & \\ \hline\hline
    B0 & RNN-T~\cite{sainath2020streaming} & - & 6.2 & 28.4 & - \\ \hline
    B1 & Transformer Rescoring~\cite{li2020parallel} & - & 5.6 & 25.9 & 2.8 \\ \hline
    B2 & LSTM Deliberation Rescoring~\cite{hu2020deliberation} & 8 & 5.5 & 24.7 & 7.7 \\ \hline
    E10 & Transformer Deliberation Rescoring & 4 & 5.2 & 23.6 & 4.8 \\ \hline
    E11 & \hspace{0.5em} + Joint Training & 4 & \textbf{5.1} & \textbf{23.0} & 4.8 \\ \hline
    E6 & Transformer Deliberation Rescoring & 1 & 5.4 & 23.9 & 3.5 \\ \hline
\end{tabular}
\caption{Comparison of RNN-T, non-deliberation transformer rescoring, LSTM-based deliberation, and transformer-based deliberation rescoring models in WERs (\%) and GFLOPS. All two-pass models are augmented with AE layers and trained with MWER loss except the RNN-T model.}
\label{tab:wer}
\end{table*}

\subsection{Comparisons}
\label{subsec:comp}
From the above analysis, we use an MWER trained 4-hypothesis transformer deliberation rescorer with additional encoder and order embedding for comparison (\texttt{E10} in Table \ref{tab:wer}). In Table \ref{tab:wer}, we compare deliberation models with a first-pass RNN-T model (\texttt{B0},~\cite{sainath2020streaming} without endpointing), LSTM deliberation rescoring (\texttt{B1})~\cite{hu2020deliberation}, and transformer rescoring (\texttt{B2},~\cite{li2020parallel} without endpointing).

First, we observe that all two-pass models perform substantially better than their first-pass RNN-T. The relative WER reduction by the proposed transformer deliberation rescorer is around 16\% for VS and 17\% for SxS. Second, comparing transformer (\texttt{E10}) and LAS (\texttt{B2}) deliberation for rescoring, we achieve 5\% relative WER reduction for VS and 4\% for SxS. Joint training (\texttt{E11}) further improves the WER reductions to 7\% relative. We also compute a second-pass GFLOPS as a sum of operations needed by both bidirectional LSTM encoder and deliberation rescorer, and show that the transformer deliberation rescorer reduces GFLOPS by 38\% relative compared to LSTM-based deliberation (from 7.7 to 4.8 GFLOPS in Table \ref{tab:wer}). Part of the reduction is because we are now using 4 hypotheses in bidirectional encoding, which reduces computation from 3.4 to 1.7 GFLOPS, for bidirectional encoding alone. The other improvement is from the transformer rescorer (4.3 to 3.1 GFLOPS, i.e., 28\% relative reduction).

We also compare our best transformer deliberation rescorer (\texttt{E11}) to a non-deliberation transformer rescorer (\texttt{B1}) in~\cite{li2020parallel}, which relies on a single attention on encoder outputs for rescoring. We achieve 9\% and 11\% relative WER reductions for VS and SxS, respectively. As for GFLOPS, our transformer deliberation rescorer has extra bidirectional encoding for first-pass hypotheses and multi-source attention, and is thus 1.7 times of \texttt{B1}. To reduce the computation of hypothesis encoding, we also present a 1-hypothesis transformer deliberation rescorer (\texttt{E6}) in Table \ref{tab:wer}. The model still performs around 4\% and 7\% relatively better than \texttt{B1}, on VS and SxS, respectively, and reduces the computation to 1.25 times. We note that there are potentially more ways to improve computation (and latency) for this model, including using a transformer encoder for hypothesis encoding and parallel rescoring as in~\cite{li2020parallel}. We will leave them as future work.

\section{Conclusion}
We presented a transformer deliberation model for rescoring. The best proposed model achieves 18\% and 19\% relative WER reductions for VS and SxS, respectively, compared to the first-pass RNN-T. The reductions are 9\% and 11\% compared to non-deliberation transformer rescoring, and 7\% compared to LSTM-based deliberation. Our transformer deliberation model is more efficient than LSTM deliberation by reducing GFLOPS by 38\% relative, where a 28\% relative reduction is due to the transformer rescorer. We show that a 1-hypothesis rescorer further reduces the computation by 27\% while maintaining some WER improvement. In future work, we will explore using transformer encoder for hypothesis encoding and parallel rescoring~\cite{li2020parallel} to improve latency.

\section{Acknowledgement}
We thank Wei Li, James Qin, and Yanzhang He for their help in computing GFLOPS and useful discussions.

\bibliographystyle{IEEEbib}
{\footnotesize\bibliography{refs}}

\end{document}